\documentclass[preprint,onecolumn,amsmath,nofootinbib]{revtex4-1}

\usepackage[utf8]{inputenc}
\usepackage{graphicx, floatrow}
\usepackage[caption=false]{subfig}
\usepackage{amssymb,amsfonts,amsmath, dsfont}
\usepackage{multirow}
\usepackage{makecell}

\usepackage{natbib}

\usepackage[]{algorithm2e}

\usepackage{hyperref}

\DeclareRobustCommand{\Fig}[1]{Figure~\ref{fig:#1}}

\DeclareRobustCommand{\Eq}[1]{Equation~\ref{eq:#1}}

\DeclareRobustCommand{\Tab}[1]{Table~\ref{tab:#1}}
\DeclareRobustCommand{\Tabs}[2]{Tables~\ref{tab:#1} and~\ref{tab:#2}}

\def \E{\textrm{E}} % expectation
\def \>{\rangle} 
\def \<{\langle}

\def\cL{\mathcal{L}}
\def\cA{\mathcal{A}}
\def\cC{\mathcal{C}}

\def\bv{\textbf{v}}

\def\bx{\textbf{x}}

\def\static{\textrm{static}}

\def\be{\begin{equation}} 
\def\ee{\end{equation}} 
\def\longrightharpoonup{\relbar\joinrel\rightharpoonup}
\def\longleftharpoondown{\leftharpoondown\joinrel\relbar}

\def\longrightleftharpoons{
  \mathop{
    \vcenter{
      \hbox{
      \ooalign{
        \raise1pt\hbox{$\longrightharpoonup\joinrel$}\crcr
	  \lower1pt\hbox{$\longleftharpoondown\joinrel$}
	  }
      }
    }
  }
}

\newcommand \bea {\begin{eqnarray}} 
\newcommand \eea {\end{eqnarray}}



\begin{document}

% 78 characters (85 allowed)
\title{Deep learning for comprehensive forecasting of Alzheimer’s Disease progression}

\author{Charles K. Fisher}
\email{drckf@unlearn.ai}
\thanks{authors listed alphabetically.}
\affiliation{Unlearn.AI, Inc., 650 California St, San Francisco, CA 94108}

\author{Aaron M. Smith}
\affiliation{Unlearn.AI, Inc., 650 California St, San Francisco, CA 94108}

\author{Jonathan R. Walsh}
\affiliation{Unlearn.AI, Inc., 650 California St, San Francisco, CA 94108}

\author{for the Coalition Against Major Diseases}
\thanks{Data used in the preparation of this article were obtained from the Coalition Against Major Diseases database (CAMD). As such, the investigators within CAMD contributed to the design and implementation of the CAMD database and/or provided data, but did not participate in the analysis of the data or the writing of this report.}

\date{\today}

% 146 words (200 allowed)
\begin{abstract}
Most approaches to machine learning from electronic health data can only predict a single endpoint. Here, we present an alternative that uses unsupervised deep learning to simulate detailed patient trajectories. We use data comprising 18-month trajectories of 44 clinical variables from 1908 patients with Mild Cognitive Impairment or Alzheimer's Disease to train a model for personalized forecasting of disease progression. We simulate synthetic patient data including the evolution of each sub-component of cognitive exams, laboratory tests, and their associations with baseline clinical characteristics, generating both predictions and their confidence intervals. Our unsupervised model predicts changes in total ADAS-Cog scores with the same accuracy as specifically trained supervised models and identifies sub-components associated with word recall as predictive of progression. The ability to simultaneously simulate dozens of patient characteristics is a crucial step towards personalized medicine for Alzheimer's Disease.
\end{abstract} 
\maketitle

% 4500 words
% 6 figures + tables
\section{Introduction}

Two patients with the same disease may present with different symptoms, progress at different rates, and respond differently to the same therapy. Understanding how to predict and manage differences between patients is the primary goal of precision medicine \cite{collins2015new}. Computational models of disease progression developed using machine learning approaches provide an attractive tool to combat such patient heterogeneity. One day these computational models may be used to guide clinical decisions; however, current applications are limited both by the availability of data and by the ability of algorithms to extract insights from those data.

Most applications of machine learning to electronic health data have used techniques from supervised learning to predict specific endpoints~\cite{rajkomar2018scalable, miotto2016deep, choi2016doctor, lasko2013computational, lipton2015learning, myers2017machine}. An alternative to developing separate supervised models to predict each characteristic is to build a single model that simultaneously predicts the evolution of many characteristics. Statistical models based on artificial neural networks (Deep Learning) provide one avenue for developing tools that can simulate patient progression in detail~\cite{choi2017generating, esteban2017real, beaulieu2017privacy}. 

Clinical data present a number of challenges that are not easily overcome with current approaches to deep learning~\cite{goldstein2017opportunities}.  For example, most clinical datasets contain multiple types of data (i.e., they are ``multimodal''), have a relatively small number of samples, and many missing observations. Dealing with these issues typically requires extensive preprocessing~\cite{miotto2016deep} or simply discarding variables that are too difficult to model. For example, one recent study focused on only four variables that were frequently measured across all 200,000 patients in an electronic health dataset from an intensive care unit~\cite{esteban2017real}. Developing methods that can overcome these limitations is a key step towards broader applications of machine learning in precision medicine.

Precision medicine is especially important for complex disorders where patients exhibit different patterns of disease progression and therapeutic responses. Alzheimer's Disease (AD) and Mild Cognitive Impairment (MCI) are complex neurodegenerative diseases with multiple cognitive and behavioral symptoms~\cite{kumar2015review}. The severity of these symptoms is usually assessed through exams such as the Alzheimer's Disease Assessment Scale (ADAS)~\cite{rosen1984new} or Mini Mental State Exam (MMSE)~\cite{folstein1975mini}. The heterogeneity of AD and related dementias makes these diseases difficult to diagnose, manage, and treat, leading to calls for better methods to forecast and monitor disease progression and to improve the design of AD clinical trials~\cite{cummings2016drug}.

A variety of disease progression models have been developed for MCI and AD using clinical data~\cite{rogers2012combining, ito2013understanding, kennedy2016post, tishchenko2016alzheimer, szalkai2017identifying} or imaging studies~\cite{mueller2005ways, risacher2009baseline, hinrichs2011predictive, ito2011disease, suk2013deep, suk2014hierarchical, liu2014early, ortiz2016ensembles, samper2017yet}.  Although previous approaches to forecasting disease progression have proven useful~\cite{corrigan2014clinical, romero2015future}, they have focused on predicting a single endpoint, such as the change in the ADAS Cognitive (ADAS-Cog) score from baseline. Given that AD is heterogeneous and multifactorial, we set out to model the progression of more than just the ADAS-Cog score. We accomplished this by simulating the progression of entire patient profiles, describing the evolution of each sub-component of the ADAS-Cog and MMSE scores, laboratory tests, and their associations with baseline clinical characteristics. 

The manuscript is structured as follows. Section \ref{ssec:processing} describes our data processing steps and Section \ref{ssec:modeling} describes our machine learning model. Section \ref{ssec:performance} assesses the goodness-of-fit of our machine learning model. Predictions for individual components are discussed in Section \ref{ssec:trajectories}. Section \ref{ssec:forecasting} assesses the accuracy of our approach, which simulates each sub-component of the cognitive scores, at predicting changes in overall disease activity measured by the ADAS-Cog exam. Finally, Section \ref{sec:discussion} discusses implications.

\section{Methods}
\label{sec:methods}

\subsection{Data Processing}
\label{ssec:processing}

Our statistical model was trained and tested on data extracted from the Coalition Against Major Diseases (CAMD) Online Data Repository for AD (CODR-AD)~\cite{romero2009coalition, neville2015development}.  The development and composition of this database have been previously described in detail \cite{neville2015development}. The CAMD database contains 6500 patients from the placebo arms of 24 clinical trials on MCI and AD.  These trials have varying duration, visit frequency, and inclusion criteria; nearly all patients have no data beyond approximately 18 months.  We chose a 3-month spacing between time points based on the visit frequency of the bulk of long-lasting patients to ensure that most patients had no gaps in their data.  The falloff in patient data after the 18-month time point led us to select that as the final time point. Therefore, patient trajectories are represented by 7 time points (0, 3, 6, 9, 12, 15, and 18 months).

Data in the CAMD database is stored in the CDISC format~\cite{kubick2007toward,hume2016current}.  The covariates used in our statistical model of AD progression originate from tables in the database on demographics, disposition events, laboratory results, medical histories, questionnaires, subject characteristics, subject visits, and vital signs.  We designated some variables, such as height, as static. Multiple values for any of the static variables were averaged to produce a single estimate. Time-dependent variables were bucketed into 90-day windows centered on each time point. Multiple entries in any window were averaged, or extremal values were taken as appropriate. Any data with units (such as laboratory tests) were converted to a common unit for each test for all patients (e.g., g/L for triglycerides). Results for both the ADAS-Cog and MMSE tests were available for many patients to the level of individual components. Individual question data were available for some patients, which we aggregated into component scores. A final processing step converted data into numerical values more suitable for statistical modeling. Categorical variables were one-hot encoded and positive continuous variables were log-transformed and standardized. All variables were transformed back to canonical form before analysis.

Our statistical model can perform imputation of missing data during training. However, using covariates that are missing in a large fraction of patients would lead to poor performance. Therefore, we chose 44 variables that were observed in a reasonably large fraction of patients. \Tabs{cognitive-variables}{non-cognitive-variables} describe each of the variables included in our analysis. Because we are interested in modeling AD progression, we focused on patients in the CAMD database with long trajectories.  This led us to select the 1908 patients from CAMD that have a valid ADAS-Cog score (i.e., data is not missing for any of the 11 components) for either of the 15-month or 18-month time points.

To summarize, we extracted 18-month longitudinal trajectories of 1908 patients with MCI or AD covering 44 variables including the individual components of the ADAS-Cog and MMSE scores, laboratory tests, and background information. The patients were randomly divided into a training group of 1335 patients, a validation group of 95 patients, and a testing group of 478 patients. The training group was only used in learning parameters of the model; the validation group was only used to monitor metrics during training, and the testing group was only used to evaluate the performance of our model. Each patient profile consisted of 44 covariates (\Tabs{cognitive-variables}{non-cognitive-variables}) that were classified as binary, ordinal, categorical, or continuous. Patient trajectories described the time evolution of all 44 variables in 3-month intervals.

\begin{table}[ht!]
\caption{Cognitive variables included in the model.}
\begin{center}
\begin{tabular}{|c|c|c|c|}
\hline
\, {\bf Category} \, & \, {\bf Name} \, & \, {\bf Type} \, & \, {\bf Temporal} \, \\
\hline 
\hline
ADAS & Commands & Ordinal & Yes \\ \cline{2-4}
ADAS & Comprehension & Ordinal & Yes \\ \cline{2-4}
ADAS & Construction & Ordinal & Yes \\ \cline{2-4}
ADAS & Delayed Word Recall & Ordinal & Yes \\ \cline{2-4}
ADAS & Ideational & Ordinal & Yes \\ \cline{2-4}
ADAS & Instructions & Ordinal & Yes \\ \cline{2-4}
ADAS & Naming & Ordinal & Yes \\ \cline{2-4}
ADAS & Orientation & Ordinal & Yes \\ \cline{2-4}
ADAS & Spoken Language & Ordinal & Yes \\ \cline{2-4}
ADAS & Word Finding & Ordinal & Yes \\ \cline{2-4}
ADAS & Word Recall & Ordinal & Yes \\ \cline{2-4}
ADAS & Word Recognition & Ordinal & Yes \\
\hline
\hline
MMSE & Attention and Calculation & Ordinal & Yes \\ \cline{2-4}
MMSE & Language & Ordinal & Yes \\ \cline{2-4}
MMSE & Orientation & Ordinal & Yes \\ \cline{2-4}
MMSE & Recall & Ordinal & Yes \\ \cline{2-4}
MMSE & Registration & Ordinal & Yes \\
\hline
\end{tabular}
\end{center}
\label{tab:cognitive-variables}
\end{table}%

\begin{table}[htp!]
\caption{Laboratory, clinical, and background variables included in the model.}
\begin{center}
\begin{tabular}{|c|c|c|c|}
\hline
\, {\bf Category} \, & \, {\bf Name} \, & \, {\bf Type} \, & \, {\bf Temporal} \, \\
\hline
\hline
Laboratory & Alanine aminotransferase & \, Continuous \, & Yes \\ \cline{2-4}
Laboratory & Alkaline phosphatase & Continuous & Yes \\ \cline{2-4}
Laboratory & Aspartate aminotransferase & Continuous & Yes \\ \cline{2-4}
Laboratory & Cholesterol & Continuous & Yes \\ \cline{2-4}
Laboratory & Creatine kinase & Continuous & Yes \\ \cline{2-4}
Laboratory & Creatinine & Continuous & Yes \\ \cline{2-4}
Laboratory & Gamma glutamyl transferase & Continuous & Yes \\ \cline{2-4}
Laboratory & Hematocrit & Continuous & Yes \\ \cline{2-4}
Laboratory & Hemoglobin & Continuous & Yes \\ \cline{2-4}
Laboratory & Hemoglobin a1c & Continuous & Yes \\ \cline{2-4}
Laboratory & Indirect bilirubin & Continuous & Yes \\ \cline{2-4}
Laboratory & Potassium & Continuous & Yes \\ \cline{2-4}
Laboratory & Sodium & Continuous & Yes \\ \cline{2-4}
Laboratory & Triglycerides & Continuous & Yes \\
\hline
\hline
Clinical & Blood pressure (diastolic) & Continuous & Yes \\ \cline{2-4}
Clinical & Blood pressure (systolic) & Continuous & Yes \\ \cline{2-4}
Clinical & Heart rate & Continuous & Yes \\ \cline{2-4}
Clinical & Weight & Continuous & Yes \\ \cline{2-4}
Clinical & Dropout & Continuous & Yes \\
\hline
\hline
Background & Age at baseline & Continuous & No \\ \cline{2-4}
Background & Geographic region & Categorical & No \\ \cline{2-4}
Background & Initial diagnosis (AD or MCI) & Binary & No \\ \cline{2-4}
Background & Past cardiovascular event & Binary & No \\ \cline{2-4}
Background & ApoE $\epsilon$4 allele count & Ordinal & No \\ \cline{2-4}
Background & Race & Categorical & No \\ \cline{2-4}
Background & Sex & Binary & No \\ \cline{2-4}
Background & Height & Continuous & No \\
\hline
\end{tabular}
\end{center}
\label{tab:non-cognitive-variables}
\end{table}%

\subsection{Machine Learning}
\label{ssec:modeling}

%%%
\begin{figure}[ht!]
\includegraphics[width=6.5in]{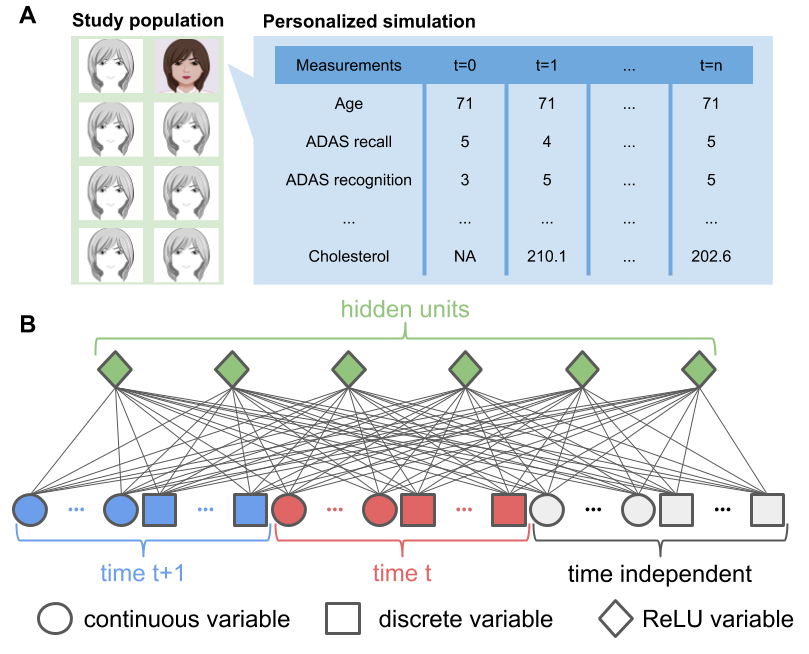}
\caption{{\bf Overview of the data and model}. A) Study data built from the CAMD database consists of 18-month longitudinal trajectories of 1908 patients with MCI or AD. Our model uses 44 variables, including the individual components of the ADAS-Cog and MMSE scores, laboratory tests, and background information. B) To capture time dependence, we model the joint distribution of the data at time $t+1$ and the data at time $t$ using a Conditional Restricted Boltzmann Machine (CRBM) with ReLU hidden units. Multimodal observations are modeled with different types of units in the visible layer and missing observations are automatically imputed.
\label{fig:schematic}}
\end{figure}
%%%

A statistical model is generative if it can be used to draw new samples from an inferred probability distribution. Generative modeling of clinical data involves two tasks: i) randomly generating patient profiles with the same statistical properties as real patient profiles and ii) simulating the evolution of these patient profiles through time. Each of these tasks is complicated by common properties of clinical data, namely that they are typically multimodal and have many missing observations. Moreover, patient progression is best regarded as a stochastic process and it is important to capture the inherent randomness of the underlying processes in order to make accurate forecasts. 

Let ${\bold x}_i(t)$ be a vector of covariates measured in patient $i$ at time $t$. Creating a generative model to solve (i) involves finding a probability distribution $P({\bold x})$ such that we can randomly draw ${\bold x}_i(t=0) \sim P({\bold x})$. Solving problem (ii) involves finding a conditional probability distribution $P({\bold x}(t) | {\bold x}(t-1))$ so that we can iteratively draw ${\bold x}_i(t) \sim P({\bold x}_i(t) | {\bold x}_i(t-1))$ to generate a patient trajectory.

Our statistical model for patient progression is a latent variable model called a Conditional Restricted Boltzmann Machine (CRBM) \cite{ackley1985learning, hinton2010practical, taylor2007modeling, mnih2011conditional}. To construct the model, the covariates were divided into two mutually exclusive subsets: {\it static} covariates that were determined solely from measurements at the beginning of the study ${\bold x}_i^{\rm static}(t=0)$, and {\it dynamic} covariates that changed during the study ${\bold x}_i^{\rm dynamic}(t)$. To train the model, we defined vectors ${\bold v}_i(t) = \{ {\bold x}_i^{\rm dynamic}(t), {\bold x}_i^{\rm dynamic}(t-1), {\bold x}_i^{\rm static}(t=0) \}$ by concatenating neighboring time points with the static covariates. All neighboring time points are combined into a single dataset used to train a single statistical model that applies to all neighboring time points. Rather than directly modeling the correlations between these covariates, a CRBM models these correlations indirectly using a vector of latent variables ${\bold h}_{\mu}(t)$. These latent variables can be interpreted in much the same way as directions identified through principal components analysis. 

The CRBM is a parametric statistical model where the probability density is defined as
\begin{equation}
p({\bold v}, {\bold h}) = Z^{-1} \exp\biggl( \sum_j a_j(v_j) + \sum_{\mu} b_{\mu} (h_{\mu}) + \sum_{j \mu} 
W_{j \mu} \frac{v_j}{\sigma_j^2} \frac{h_{\mu}}{\epsilon_{\mu}^2} \biggr) \,,
\end{equation}
and $Z$ is a normalization constant that ensures the total probability integrates to one. Here, $a_j(v_j)$ and  and $b_{\mu}(h_{\mu})$ are functions that characterize the data types of covariate $v_j$ and latent variable $h_{\mu}$, respectively. The parameters $\sigma_j$ and $\epsilon_{\mu}$ set the scales of $v_j$ and $h_{\mu}$, respectively. We used 50 normally distributed latent variables that were lower truncated at zero, which is known as a rectified linear (ReLU) activation function in the machine learning literature \cite{tubiana2017emergence}. To deal with missing data, we divide the visible vector ${\bold v}$ into mutually exclusive groups ${\bold v}_{missing}$ and ${\bold v}_{observed}$ and impute the missing values by drawing from the conditional distribution $p({\bold v}_{missing} \, \vert \, {\bold v}_{observed})$. 

Traditionally, CRBMs are trained to maximize the likelihood of the data under the model using stochastic maximum likelihood \cite{tieleman2008training}. Recent results have shown that one can improve on maximum likelihood training of RBMs by adding an additional term to the loss function that measures how easy it is to distinguish patient profiles generated from the statistical model from real patient profiles \cite{fisher2018boltzmann}. Therefore, we used a combined maximum likelihood and adversarial training method to fit the CRBM; more details of the machine learning methods are described in the Supporting Information. An overview of our statistical model is depicted in \Fig{schematic}. 

We generated two types of synthetic patient trajectories with a CRBM: i) synthetic trajectories starting from baseline values for real patients, and ii) entirely synthetic patients.  The first type is useful for many tasks in precision medicine and clinical trial simulation, while the second type has interesting applications for maintaining the privacy of clinical data~\cite{dankar2013practicing}.  To generate trajectories of type (i), an initial population of patients was selected and then the model was used to predict their future state.  To accomplish this, we started with baseline data and used the CRBM to iteratively add new time points. To generate trajectories of type (ii), entirely synthetic patients were generated by first simulating the baseline data, then iteratively adding new time points so that the patient data was entirely simulated.

\section{Results}
\label{sec:results}

%%%
\begin{figure}[t!]
\includegraphics[width=6.5in]{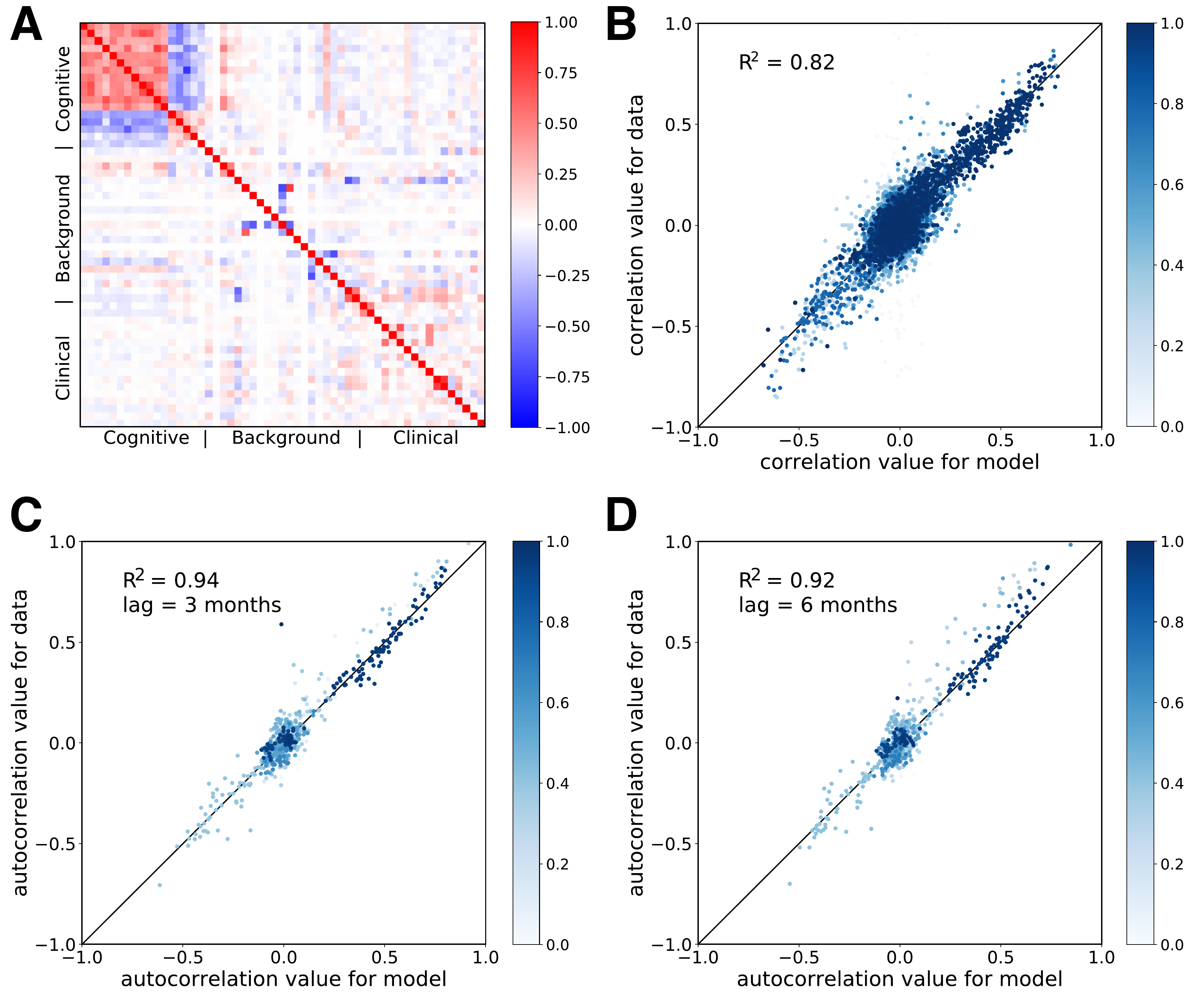} 
\caption{{\bf The model has good generative performance.} A) Correlations between variables as predicted by the model (below the diagonal) and calculated from the data (above the diagonal). Components of the cognitive scores are strongly correlated with each other, but not with other clinical data. B) Scatterplot of observed vs predicted correlations for each time point, over all times. C) Scatterplot of observed vs predicted autocorrelations with time lag of 3 months. D) Scatterplot of observed vs predicted autocorrelations with time lag of 6 months. The color gradient in B -- D represents the fraction of observations where the variables used to compute the correlation were present; lighter colors mean more of the data was missing.  The $R^2$ values shown are from a least squares fit weighted by this fraction (of data present when computing the correlations).
\label{fig:fit}}
\end{figure}
%%%

\subsection{General model performance}
\label{ssec:performance}

As an initial measure of performance, we assessed the ability of the CRBM to generate marginal distributions of each variable using entirely synthetic patients.  The CRBM generated time series that accurately captures the marginal distributions of cognitive exam scores, laboratory tests, and clinical data of real AD patients (Supporting Information). Beyond marginal distributions, equal-time and lagged autocorrelations are more important factors in forecasting disease progression, so we assessed the CRBM's ability to model these data.  Entirely synthetic patient trajectories have correlations that model the data well (\Fig{fit}A). The variables can be reasonably grouped into three categories: cognitive scores, laboratory and clinical tests, and background information. There are strong correlations between variables belonging to the same category but only weak inter-category correlations (\Fig{fit}A). Unfortunately, this also implies that neither the laboratory tests nor background variables are strongly correlated with the primary clinical endpoints captured by cognitive assessments for AD. Even though the CRBM only incorporates a direct connection between neighboring time points, this is sufficient to reproduce equal-time and lagged autocorrelations between variables (\Fig{fit}B-D), even for time lags greater than 3 months.  Collectively, these results suggest the model has excellent generative performance.

\subsection{Simulating conditional patient trajectories}
\label{ssec:trajectories}

%%%
\begin{figure}[thp!]
\includegraphics[width=3.5in]{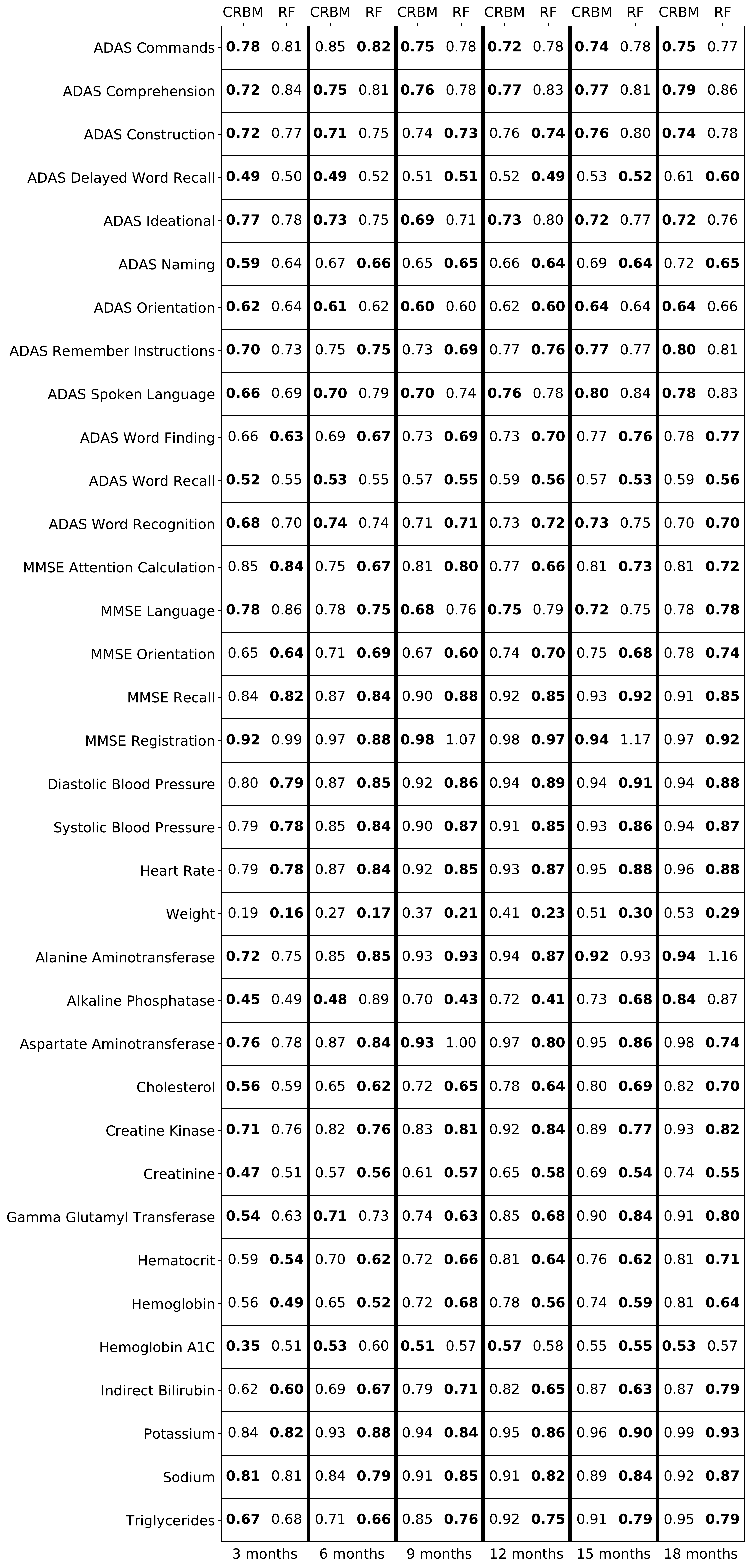} 
\caption{{\bf The model accurately forecasts across variables.} Relative errors of the model (CRBM) and a random forest (RF) specifically trained to predict the value of a single variable at a single time point.  The root mean square (RMS) errors are scaled by the standard deviation of the data to be predicted.  Predictions are shown for every time-dependent variable except dropout.  At each time point and for each variable, the better of the random forest and CRBM predictions is shown in bold.
\label{fig:trajectories}}
\end{figure}
%%%

Predictions for any unobserved characteristics of a patient can be computed from our model by generating samples from the model distribution conditioned on the values of all observed variables. Sampling from the conditional distributions can be used to fill-in any missing observations (i.e., imputation) or to forecast a patient's future state. The ability to sample from any conditional distribution is one advantage a modeling framework based on CRBMs has over alternative generative models based on directed neural networks. 

For each patient in the test set, we computed a forecast for their entire trajectory conditioned on their baseline covariates. That is, we used the CRBM to numerically compute the conditional expected value $\E[\bold{x}_i(t) | \bold{x}_i(t=0)]$. Next, we evaluated the root mean square (RMS) error of the CRBM predictions on each variable at each time point past baseline. For comparison, we trained a series of Random Forest (RF) models that use the baseline data to predict each of the 35 time-dependent variables for all 6 time points. Note that there is a separate RF model for each variable at each time point -- a total of 210 different RF models. We also trained an ensemble of 6 multivariate RFs -- each one predicted all 35 covariates for a given time point -- but were unable to get reasonable accuracies (see Supporting Information). The RMS error of the random forest prediction sets a benchmark for a predictive model that is specially trained for an individual problem. By contrast, a single CRBM model is used to predict all variables, and all time points. \Fig{trajectories} presents a detailed comparison between the single CRBM and the ensemble of 210 RF models. The accuracy of the CRBM is close to the specialized RF model for each variable and time point, with the CRBM generally outperforming the RF on the components of ADAS-Cog. 

One can think of the ensemble of RF models as approximating the predictions of a factorized generative model. That is, one could construct a simpler probabilistic model by assuming that the variables are independent when conditioned on the baseline values, i.e. $p(\bold{x}(t) | \bold{x}(t=0)) \approx \prod_j p(x_j(t) | \bold{x}(t=0))$. While this factorized model can make accurate predictions for individual variables in isolation, it cannot generate realistic trajectories that capture the correlations between the covariates (which will be zero by construction). By contrast, the CRBM achieves equivalent accuracy on individual prediction problems while also correctly modeling the correlation structure. More details on the comparison between RFs and the CRBM are provided in the Supporting Information.

In summary, stochastic simulations of disease progression have two main advantages compared to supervised machine learning models that aim to predict a single, predefined endpoint. The first is the simultaneous modeling of entire patient profiles in a way that correctly captures correlations between covariates. This allows for the quantitative exploration of alternative endpoints and different patient subgroups. The second is that stochastic simulations provide in-depth estimates of risk for individual patients that can be aggregated to estimate risks in larger patient populations.  Moreover, our model provides accurate estimates of its uncertainty in addition to forecasts for expected progression of individual patients (Figures S1 and S8). Patient heterogeneity manifests in more complex ways than just a shift in the mean outcome -- there are changes in the variance, skew, and shape of the distribution of model predictions for each patient. Personalized approaches to AD therapy will have to predict and address these different types of risk. The ability to simultaneously compute predictions and confidence intervals for multiple characteristics of a patient is a key feature of our approach and an important step towards comprehensive simulations of disease progression.

\subsection{Forecasting and interpreting disease progression}
\label{ssec:forecasting}

%%%
\begin{figure}[thp!]
\includegraphics[width=6.5in]{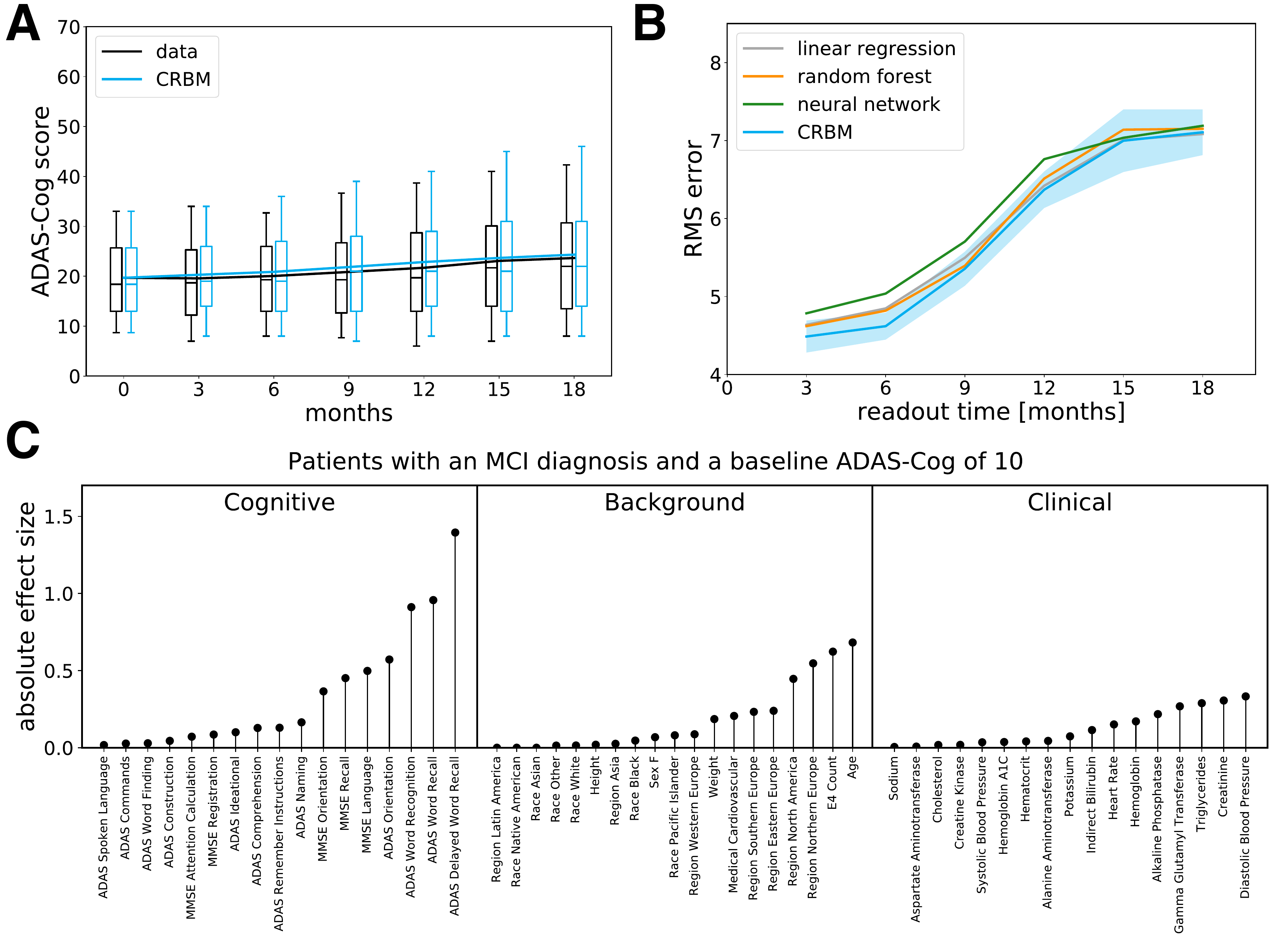} 
\caption{{\bf The model accurately forecasts progression and allows for interpretation.} A) Box plot of the ADAS-Cog score over time computed from the data and the model. The line shows the mean, and the whiskers show the 10$^{\rm th}$ and 90$^{\rm th}$ percentiles.  B) Out-of-sample predictive accuracy for the change in ADAS-Cog score from baseline (i.e., $t=0$) for different study durations. Separate neural network, random forest, and linear regression models were trained to predict the change in ADAS-Cog score from baseline for each study duration.  The blue band shows the uncertainty on the CRBM prediction. C) We created a simulated patient population with MCI and an initial ADAS-Cog score of 10, and simulated the evolution of each synthetic patient for 18 months. The 5\% of synthetic patients with the largest ADAS-Cog score increase were designated ``fast progressors'' and the bottom 5\% of patients with the smallest ADAS-Cog score increase were designated ``slow progressors''. Differences between the fast and slow progressors (the ``absolute effect size'') were quantified using the absolute value of Cohen's $d$-statistic, which measures the mean difference divided by a pooled standard deviation~\cite{cohen1988statistical}.
\label{fig:progression}}
\end{figure}
%%%

We now turn to disease progression to evaluate the CRBM and understand how it provides interpretative power for MCI and AD.  Our model is trained to simulate the evolution of the individual components of the cognitive exams, laboratory tests, and clinical data. As a result, it is also possible to simulate the evolution of any combination of these variables, such as the 11-component ADAS-Cog score that is commonly used as a measure of overall disease activity.  Note that the ADAS delayed word recall component, which is present in the dataset, is not part of the 11-component ADAS-Cog score but can be used as an additional probe of disease severity, especially for MCI~\cite{sano2011adding}.  \Fig{progression}A shows a box plot describing the evolution of the ADAS-Cog score distribution within the population. The data and model show the same trend -- an increase in the mean ADAS-Cog score with time along with a widening right tail of the distribution. This implies that much of the trend of increasing ADAS-Cog scores in the population is driven by a subset of patients. 

Simulations from the model can be run for each individual patient in order to forecast their disease progression. Despite only being trained on data with a 3-month time lag, the model makes accurate predictions out to at least 18 months (\Fig{progression}B). In \Fig{progression}B, we have compared the accuracy of the CRBM predictions for the change in ADAS-Cog score from baseline to each possible endpoint in 3-month steps through 18 months to a variety of supervised models (a linear regression, a random forest, and a deep neural network). Each of the supervised models was trained to predict a specific endpoint (e.g., the change in ADAS-Cog score after 6 months). The CRBM is the best performing model, though the accuracies of the four types of predictors converge for time periods of 15 months or longer. The strong relative performance of the CRBM on this task is remarkable given that (i) it was only trained to perform 3-month ahead simulations and (ii) it was not directly trained to predict the aggregate ADAS-Cog score. More details on the comparison are provided in the Supporting Information.

To gain more insight into the origin of fast and slow progressing patients, we simulated 18-month patient trajectories conditioned on a baseline ADAS-Cog score of 10 and an initial diagnosis of MCI. This initial ADAS-Cog score was chosen because it is representative of a typical patient with MCI. The 5\% of synthetic patients with the largest ADAS-Cog score increase were designated ``fast progressors'' and the bottom 5\% of synthetic patients with the smallest ADAS-Cog score increase were designated ``slow progressors''. Differences between the fast and slow progressors (the ``absolute effect size'') were quantified using the absolute value of Cohen's $d$-statistic~\cite{cohen1988statistical}, as shown in \Fig{progression}C. The majority of baseline variables are not associated with disease progression; however, there are strong associations with cognitive tests based on recall (i.e., MMSE recall, ADAS word recall, and ADAS delayed word recall) and word recognition. That is, patients with poor performance on the ADAS delayed word recall test tend to progress more rapidly -- even after controlling for the total ADAS-Cog score. Variables associated with progression in patients who already have AD are described in the Supporting Information.

\section{Discussion}
\label{sec:discussion}

The ability to simulate the stochastic disease progression of individual patients in high resolution could have a transformative impact on patient care by enabling personalized data-driven medicine. Each patient with a given diagnosis has unique risks and a unique response to therapy. Due to this heterogeneity, predictive models cannot currently make individual-level forecasts with a high degree of confidence. Therefore, it is critical that data-driven approaches to personalized medicine and clinical decision support provide estimates of their uncertainty in addition to expected outcomes.

Previous efforts for modeling disease progression in AD have focused on predicting changes in predefined outcomes such as the ADAS-Cog score or the probability of conversion from MCI to AD ~\cite{rogers2012combining, ito2013understanding, kennedy2016post, tishchenko2016alzheimer, szalkai2017identifying, mueller2005ways, risacher2009baseline, hinrichs2011predictive, ito2011disease, suk2013deep, suk2014hierarchical, liu2014early, ortiz2016ensembles, samper2017yet}. Here, we have demonstrated that an approach based on unsupervised deep learning can create stochastic simulations of entire patient trajectories that achieve the same level of performance on individual prediction tasks as specific models while also accurately capturing correlations between variables. Deep learning-based generative models provide much more information than specific models, thereby enabling a simultaneous and detailed assessment of different risks. 

Our approach to modeling patient trajectories in AD overcomes many of the limitations of previous applications of deep learning to clinical data \cite{goldstein2017opportunities, miotto2016deep, esteban2017real, choi2017generating}. CRBMs can directly integrate multimodal data with both continuous and discrete variables, and time-dependent and static variables, within a single model. In addition, bidirectional models like CRBMs can easily handle missing observations in the training set by performing automated imputation during training. Combined, these factors dramatically reduce the amount of data preprocessing steps needed to train a generative model to produce synthetic clinical data. We found that a single time-lagged connection was sufficient for explaining temporal correlations in AD; additional connections may be required for diseases with more complex temporal evolution. 

The utility of cognitive scores as a measure of disease activity for patients with AD has been called into question numerous times~\cite{benge2009well}. Here, we found that the components of the ADAS-Cog and MMSE scores were only weakly correlated with other clinical variables. One possible explanation is that the observed stochasticity may simply reflect heterogeneity in performance on the cognitive exam that cannot be predicted from any baseline measurements. However, we did find that some of the individual components of the baseline cognitive scores are predictive of progression. Specifically, patients with poor performance on word recall tests tend to progress more rapidly than other patients, even after controlling for the ADAS-Cog score. 

\section{Conclusions}

This work provides a proof-of-concept that patient-level simulations are technologically feasible with the right tools and data. Nevertheless, there are a number of improvements to our dataset and methodology that are important steps for future research. Here, we limited ourselves to modeling 44 variables that are commonly measured in AD clinical trials. We excluded some interesting covariates such as Leukocyte populations because they were not measured in the majority of patients in our dataset constructed from the CAMD database. We also lack data from neuroimaging studies and tests for levels of amyloid-$\beta$. Incorporating additional data into our model development will be a crucial next step, especially as surrogate biomarkers become a standard part of clinical trials. 

The approach to simulating disease progression that we describe here can be easily extended to other diseases. Widespread application of deep generative models to clinical data could produce synthetic datasets with lower privacy concerns than real medical data~\cite{beaulieu2017privacy}, or could be used to run simulated clinical trials to optimize study design or as synthetic control arms. In certain disease areas, tools that use simulations to forecast risks for specific individuals could help doctors choose the right treatments for their patients. Currently, progress towards these goals is slowed by the limited availability of high quality longitudinal health datasets and the limited ability of current machine learning methods to produce insights from these datasets. 

\section{Acknowledgements}

We would like to thank Yannick Pouliot, Pankaj Mehta, and Diane Dickel for helpful comments while preparing the manuscript. Data used in the preparation of this article were obtained from the Coalition Against Major Diseases (CAMD) database. In 2008, Critical Path Institute, in collaboration with the Engelberg Center for Health Care Reform at the Brookings Institution, formed the Coalition Against Major Diseases (CAMD). The Coalition brings together patient groups, biopharmaceutical companies, and scientists from academia, the U.S. Food and Drug Administration (FDA), the European Medicines Agency (EMA), the National Institute of Neurological Disorders and Stroke (NINDS), and the National Institute on Aging (NIA). The Coalition Against Major Diseases (CAMD) includes over 200 scientists from member and non-member organizations. The data available in the CAMD database has been volunteered by CAMD member companies and non-member organizations.

\bibliography{camd}

\onecolumngrid

\section{Supporting Information}

\subsection{Data Processing
\label{app:data_processing}}

The CAMD database stores data using CDISC standards, specifically the Study Data Tabulation Model (SDTM), which defines a common schema for clinical trial data and is the required standard for clinical data submissions to the United States Food and Drug Administration (FDA).  In this format the data is already highly structured; therefore it is possible to develop data processing pipelines that can apply to SDTM data in general and not simply the particular database used here.  We describe the general architecture of our data processing pipeline and the CAMD-specific processing used.

The goal of our processing pipeline is to arrive at data that may be directly used by machine learning algorithms to build patient-level models.  This means:
\begin{itemize}
\item Data must be numerically formatted, such as numeric values, ordinal values for scores, and one-hot encoding for categorical variables.  For text or image data, this may involve feature extraction, e.g. through a word2vec model or an autoencoder.
\item Data must be patient-specific, and can extend over time in regular intervals.  For example, if we have cholesterol measurements for a given patient at 1, 2, 5, and 12 months, but are modeling the population at 3-month intervals, then we may average the 1- and 2-month time point values and will have a missing entry between the 5- and 12-month values.  
\end{itemize}

Data arrives in Comma Separated Value (CSV) formatted text files, with abbreviation encodings for file and variable names.  Many of these abbreviation are generic to SDTM and some apply specifically to disease areas.  A translation table, such as one provided by CAMD, may be used to automatically convert abbreviations to human-readable names.  Using this translation and simple type inference on variables, the data is ingested into a SQL database via a simple script.  We label this data in this form as the {\it raw database}.

The main component of the processing pipeline extracts data appropriate for training and evaluating machine learning algorithms.  This is done on a per-variable basis, meaning the primary functions in the pipeline produce data for only a single variable; this processing is then repeated over all variables of interest.  Processed data is stored in the {\it processed database} and may directly be used to construct datasets for machine learning.  The steps in the processing are:
\begin{itemize}
\item Declare which columns and tables from the raw database will be used to produce the data for a given variable.
\item Declare a processing function to convert this data into the appropriate form.
\item Declare a location in the processed database where the data will be stored.
\item Query the raw database for the data, apply the processing function, and store the result in the processed database.
\end{itemize}
The processing functions may be common, such as one-hot encoding categorical labels, or they may be custom, such as standardizing units for a particular laboratory measurement.  Such custom functions form the bulk of database-specific code that must be written.  All of the above processing steps can easily be encoded in configuration files, meaning the process of preparing data for machine learning is simple and repeatable.

Finally, datasets may be constructed from the processed database by merely specifying which variables are to be used.  This step is also performed via a configuration file.  Additional filtering of patients, e.g. by requiring they have data present for a certain number of time points, is straightforward to apply.

We have developed a {\tt python} library to process data as described above.  This library fully handles the interface with the SQL database, has common data conversion functions, and provides utilities to provide summary statistics and type inference for variables in a dataset.  For any specific project, such as the CAMD database, most of the processing is set up by writing {\tt YAML} configuration files that are simple, human-readable, and easily verified.  The remainder involves writing custom processing functions in {\tt python} for specific variables.  This setup makes it straightforward to apply our machine learning models to other clinical data modeling problems.

\subsubsection{Variables Used in Training
\label{app:variables_details}}

Variables relevant to modeling AD progression were extracted from the CAMD database using the method described above, and 44 variables without substantial missing data were identified and used for the model.  \Tabs{cognitive-variables-specific}{non-cognitive-variables-specific} lists all variables, their units, and specific processing considerations for each.  Each laboratory test variable is converted to the units given, and all transformations applied to train the models are inverted for analysis.

\begin{table}[t!]
\caption{Cognitive variables included in the model.}
\begin{center}
\begin{tabular}{|c|c|c|c|}
\hline

\, {\bf Category} \, & \, {\bf Name} \, & \, {\bf Units} \, & \, {\bf Notes} \, \\
\hline 
\hline
ADAS & Commands & counts & $0-5$ ordinal range \\ \cline{2-4}
ADAS & Comprehension & counts & $0-5$ ordinal range \\ \cline{2-4}
ADAS & Construction & counts & $0-5$ ordinal range \\ \cline{2-4}
ADAS & Delayed Word Recall & counts & $0-10$ ordinal range \\ \cline{2-4}
ADAS & Ideational & counts & $0-5$ ordinal range \\ \cline{2-4}
ADAS & Instructions & counts & $0-5$ ordinal range \\ \cline{2-4}
ADAS & Naming & counts & $0-5$ ordinal range \\ \cline{2-4}
ADAS & Orientation & counts & $0-8$ ordinal range \\ \cline{2-4}
ADAS & Spoken Language & counts & $0-5$ ordinal range \\ \cline{2-4}
ADAS & Word Finding & counts & $0-5$ ordinal range \\ \cline{2-4}
ADAS & Word Recall & counts & $0-10$ ordinal range \\ \cline{2-4}
ADAS & Word Recognition & counts & \, $0-12$ ordinal range \, \\
\hline
\hline
MMSE & Attention and Calculation & counts & $0-5$ ordinal range \\ \cline{2-4}
MMSE & Language & counts & $0-9$ ordinal range \\ \cline{2-4}
MMSE & Orientation & counts & $0-10$ ordinal range \\ \cline{2-4}
MMSE & Recall & counts & $0-3$ ordinal range \\ \cline{2-4}
MMSE & Registration & counts & $0-3$ ordinal range \\
\hline
\end{tabular}
\end{center}
\label{tab:cognitive-variables-specific}
\end{table}%

\begin{table}[ht!]
\caption{Laboratory, clinical, and background variables included in the model.}
\begin{center}
\begin{tabular}{|c|c|c|c|}
\hline
\, {\bf Category} \, & \, {\bf Name} \, & \, {\bf Units} \, & \, {\bf Notes} \, \\
\hline
\hline
Laboratory & Alanine aminotransferase & $\mu$kat/l & log-standarized for training \\ \cline{2-4}
Laboratory & Alkaline phosphatase & $\mu$kat/l & log-standarized for training \\ \cline{2-4}
Laboratory & Aspartate aminotransferase & $\mu$kat/l & log-standarized for training \\ \cline{2-4}
Laboratory & Cholesterol & mmol/l & log-standarized for training \\ \cline{2-4}
Laboratory & Creatine kinase & iu/cl & log-standarized for training \\ \cline{2-4}
Laboratory & Creatinine & mg/dl & log-standarized for training \\ \cline{2-4}
Laboratory & Gamma glutamyl transferase & iu/dl & log-standarized for training \\ \cline{2-4}
Laboratory & Hematocrit & counts & log-standarized for training \\ \cline{2-4}
Laboratory & Hemoglobin & g/dl & log-standarized for training \\ \cline{2-4}
Laboratory & Hemoglobin a1c & \% & log-standarized for training \\ \cline{2-4}
Laboratory & Indirect bilirubin & mg/dl & log-standarized for training \\ \cline{2-4}
Laboratory & Potassium & mmol/l & log-standarized for training \\ \cline{2-4}
Laboratory & Sodium & mmol/cl & log-standarized for training \\ \cline{2-4}
Laboratory & Triglycerides & g/l & log-standarized for training \\
\hline
\hline
Clinical & Blood pressure (diastolic) & mmHg & log-standarized for training \\ \cline{2-4}
Clinical & Blood pressure (systolic) & mmHg & log-standarized for training \\ \cline{2-4}
Clinical & Heart rate & bpm & log-standarized for training \\ \cline{2-4}
Clinical & Weight & kg & log-standarized for training \\ \cline{2-4}
Clinical & Dropout & - & \, 1 for dropout before the next time \, \\
\hline
\hline
Background & Age at baseline & Years & {\tt `>89'} $\to 90$, log-std$^{\rm ized}$ for training \\ \cline{2-4}
Background & Geographic region & - & 1-hot, 7 labels built from country \\ \cline{2-4}
Background & Initial diagnosis (AD or MCI) & - & Bernoulli \\ \cline{2-4}
Background & Past cardiovascular event & - & Bernoulli \\ \cline{2-4}
Background & ApoE $\epsilon$4 allele count & counts & 0, 1, or 2 \\ \cline{2-4}
Background & Race & - & 1-hot, 6 labels \\ \cline{2-4}
Background & Sex & - & Bernoulli, 1 if female \\ \cline{2-4}
Background & Height & cm & log-standarized for training \\
\hline
\end{tabular}
\end{center}
\label{tab:non-cognitive-variables-specific}
\end{table}%

Of the 6500 patients in the CAMD database, very few have data after approximately 18 months from baseline.  A 3-month (90-day) interval was a suitable interval such that most patients have data at every time point; shorter intervals yielded groups of patients without data at some time points.  Therefore, we chose to represent all temporal variables in 3-month intervals from 0 (baseline) to 18 months, giving 7 available time points for each patient.

\subsubsection{Patients Used in Training
\label{app:patient_details}}

To model progression, we are most interested in patients with longer trajectories.  Therefore, we selected the 1908 patients that have data at the 15- or 18-month time points.  These patients were randomly divided into three groups: training (1335 patients, or 70\%), validation (95 patients, or 5\%), and testing (478 patients, or 25\%).  The CRBM is trained only on the training group, with the validation group used to evaluate the model's performance during training.  The supervised models predicting progression are trained on the combination of the training and validation groups using 5-fold nested cross validation.  Analysis is performed only on the testing group.

\subsection{Motivation for CRBMs
\label{app:crbm_details}}

Boltzmann machines are a well-known, standard machine learning algorithm for modeling relationships between data.  They provide several features critical to modeling clinical data not found in most machine learning models:
\begin{itemize}
\item They can easily model multimodal data.  Different neuron types may be used to model continuous numeric, ordinal, Bernoulli, or categorical data.
\item They allow for conditional and generative sampling.  If some clinical data is known for a patient, it can be used to predict unknown data for that patient.  For example, an initial population of an AD clinical trial may be defined in terms of standard inclusion criteria, such as age, sex ratio, and ADAS-Cog scores, and the remaining baseline data and any future data may be predicted.
\item As a consequence, they naturally handle missing data.  The model itself may be used to impute missing values from the learned joint probability distribution of the data.  This may be done {\it during training}, meaning missing data can be directly fed into the model.
\item They are stochastic, meaning data may be sampled.  Stochastic models naturally provide an estimate of their uncertainty through this sampling.  For clinical data, the consequence is that the model returns both a prediction and an uncertainty for any clinical variable being predicted.
\end{itemize}

Conditional Restricted Boltzmann Machines \cite{ackley1985learning, hinton2010practical, taylor2007modeling, mnih2011conditional} provide a way to model time series data using the natural capabilities of Boltzmann machines.  Our CRBM contains the visible units for multiple time points, with a standard hidden layer.  The visible units are organized as:
%%%
\be
\bv_{\rm CRBM} = \bv_\static \oplus \bv_t \oplus \cdots \oplus \bv_{t+k} \,,
\ee
%%%
where $k$ is the time lag of the model.  In our model, we use $k=1$, so that two time points are learned simultaneously.  The static units are only used once, as they are constant over all times.  The model learns the complete joint probability distribution between all $k+1$ adjacent time points simultaneously.  That means that {\emph any} conditional sampling of the data may be performed, such as predicting the data for a time point given the previous $k$ time points.  A baseline cohort may be simulated by sampling from the model and using the first time point.  This treatment of the data to allow for learning inter-dependence between time points is the only distinction of a CRBM over a standard RBM.

\subsection{Details of Training
\label{app:training_details}}

The CRBM is trained on the data from adjacent pairs of time points.  If $\bx_t$ is the vector of time-dependent variables for a patient at time $t$ and $\bx_\static$ is the vector of static variables for the same patient, then the visible units used to train the CRBM are $\bv = \{\bx_{t+1}, \bx_t, \bx_\static\}$, a concatenation of the data from the adjacent time points $t$ and $t+1$ with the static variables represented only once.

When training the CRBM, the data for each patient in the training and validation groups are reorganized into all adjacent pairs of frames.  Since each patient has data for 7 time points, they contribute 6 pairs of time points (which we will call samples). Inside of each group samples are all shuffled so that minibatches contain a mixture of patients and times.

The CRBM has a single hidden layer of 50 ReLU units, and is trained using the methods described in~\cite{fisher2018boltzmann}.  The objective function $\cC$ is a linear combination of log-likelihood $\cL$ and adversarial $\cA$ objectives,
%%%
\be \label{eq:CRBMobj}
\cC =  -\gamma \cL - (1-\gamma) \cA \,,
\ee
%%%
where $\gamma$ is a parameter weighing the relative size of the two objectives.  Parameters used to train the CRBM are listed in \Tab{training_params}.  The training setup is similar to those used in~\cite{fisher2018boltzmann}; here we use a random forest classifier for the adversary.  Temperature-driven sampling, where the temperature is sampled from an autocorrelated Gamma distribution, is not used, though the model performance is not especially sensitive to this choice.

Notable dynamics were observed during the training process.  Within 100 epochs, metrics monitored during training such as KL divergence and reverse KL divergence achieve values close to their final values.  Sampling from the model at this stage, the model has relatively poorer performance for patients with extremal ADAS-Cog scores than those near the mode.  Most importantly, during the early stages of training the model has a strong regression to the mean effect for ADAS-Cog score outliers, where the model predicts patients with a low score progress rapidly and those with a high score {\it improve} -- the opposite of what is observed in the data.  Continuing training allows the model to unlearn this behavior and correctly learn a progression from the mean, where higher scoring individuals progress more rapidly.  We expect that this feature of training, where it takes a longer time to effectively learn the behavior of outliers, is common.

\begin{table}[t!]
\caption{Hyperparameters used to train the CRBM.}
\begin{center}
\begin{tabular}{|c|c|}
\hline
\, {\bf Hyperparameter} \, & \, {\bf Value / Notes} \, \\
\hline 
\hline
number of epochs & 2000 \\
\hline
batch size & 100 \\
\hline
training/validation fractions & $95\% \, / \, 5\%$ \\
\hline
learning rate & \, 0.005 initial; 0 final; linear decay \, \\
\hline
optimizer & ADAM, beta $(0.9, 0.999)$ \\
\hline
Monte Carlo steps (sampling) & 50 \\
\hline
Monte Carlo steps (imputation) & 2 \\
\hline
driven sampling $\beta$ & 0 \\
\hline
\, likelihood weight $\gamma$ (\Eq{CRBMobj}) \, & 0.3 \\
\hline
adversary & random forest, 5 trees with max depth 5 \\
\hline
\end{tabular}
\end{center}
\label{tab:training_params}
\end{table}%

\subsubsection{Details of Progression Predictions
\label{app:supervised_details}}

This subsection gives details on the modeling for Figure 4B.  A linear regression, random forest regression, and neural network were trained to predict the ADAS-Cog score change from the baseline patient data at a single given readout time.  The CRBM is also used to predict the same score change. The performance of all the algorithms are very similar; it is likely additional patients or additional data predictive of patient progression would be needed to substantially improve the performance of the models.  For example, we found that the addition of the ApoE $\epsilon4$ allele count to the baseline variables decreases the RMS error by 5--10\%.  This section gives details on the training of the supervised models and the evaluation of all models.

The supervised models are trained from the baseline time point data to predict the ADAS-Cog score change from baseline to readout.  All possible time points (3, 6, 9, 12, 15, and 18 months) are used as readout times, and separate supervised models must be trained for each readout time.  The same CRBM model may be used for all readout times.  

However, there is missing data for many patients.  We exclude any patients that having any missing ADAS-Cog components at baseline or readout, ensuring that valid labels can be defined.  We mean impute other missing baseline variables from the training data.  After this screening, the number of training patients for the supervised algorithms and testing patients for all algorithms as a function of readout time is given in \Tab{readout_counts}.

\begin{table}[t!]
\caption{Number of training and testing patients used to predict ADAS-Cog score progression as a function of readout time.}
\begin{center}
\begin{tabular}{|c|c|c|}
\hline
\, {\bf Readout Time [months]} \, & \, {\bf Training Patients} \, & \, {\bf Testing Patients} \, \\
\hline 
\hline
3 & 1404 & 468 \\
\hline
6 & 1401 & 467 \\
\hline
9 & 1392 & 461 \\
\hline
12 & 1392 & 461 \\
\hline
15 & 1402 & 468 \\
\hline
18 & 1285 & 439 \\
\hline
\end{tabular}
\end{center}
\label{tab:readout_counts}
\end{table}%

The supervised algorithms are trained using 5-fold nested cross validation, drawing samples from the same dataset as the training and validation samples for the CRBM.  For the neural network, only one set of hyperparameters was chosen and so there was only a single cross validation loop.  All algorithms are evaluated on the same test set.  \Tab{supervised_architectures} gives the architectures and hyperparameters for each of the supervised algorithms.  Note that for the supervised algorithms, a new model was trained for every timepoint, while the CRBM is the same model over all timepoints.

\begin{table}[t!]
\caption{Supervised models used to predict ADAS-Cog score progression.}
\begin{center}
\begin{tabular}{|c|c|c|}
\hline
\, {\bf Model} \, & \, {\bf Architecture} \, & \, {\bf Hyperparameters} \, \\
\hline 
\hline
Linear Regression & ridge ($L_2$ regularization) & $\alpha \in \{10^k\}_{k=-3}^{2}$   \\
\hline
Random Forest & 100 trees & max depth $\in \{2^k\}_{k=2}^6$ \\
\hline
Neural Network & \makecell{ \, 2 hidden layers (30, 10) units \, \vspace{-0.4em} \\ ReLU activations} & \makecell{\, ADAM learning rate = 0.02 \, \vspace{-0.4em} \\ batch size = 25 \vspace{-0.4em} \\ 20 epochs} \\
\hline
\end{tabular}
\end{center}
\label{tab:supervised_architectures}
\end{table}%

Once trained, the supervised algorithms are used to predict the ADAS-Cog score change for the test data, and a root mean square (RMS) error over the test set is computed.  This is done for each readout time.  The predictions for the CRBM (over all readout times) are obtained by repeatedly simulating patient trajectories from the baseline time point.  For each simulation, the ADAS-Cog score change is recorded, yielding a distribution of score changes {\it for each patient} that represents the probabilistic distribution of predictions made by the CRBM.  The mean of this distribution is the prediction of the CRBM for the given patient.  The RMS error of these predictions is computed, as well as its standard error.  These results make up the data shown in Figure 4B.

\subsubsection{Details of Trajectory Progression Predictions
\label{app:trajectory_supervised_details}}

This subsection gives details on the modeling for Figure 3.  The approach is the natural extension of the methodology described in the previous subsection.  

A random forest is trained to predict a single time-dependent variable at a single readout time, meaning over all 35 time-dependent variables and all 6 possible readout times, 210 different random forest models are trained.  The input data are the baseline variables, with mean imputation used in the case of missing data.  Samples where either the baseline value or the readout value (the label) are missing are excluded.  The root mean square (RMS) error is computed over the test data, again only using samples where the variable being modeled is present at both baseline and the readout time.  For the CRBM, predictions are made by repeatedly simulating patients conditioned on their baseline data, and taking the mean for each patient as the CRBM prediction.  The RMS error can then be computed, using the same test samples on which the random forest models were evaluated.  

It is helpful to normalize these errors by the standard deviation of the value to be predicted.  An error ratio of 1 implies that the prediction is no better than predicting the mean of the test data, and an error ratio well below 1 implies that the prediction is highly precise at a per-patient level.

\subsection{Additional Results
\label{app:additional_results}}

There are many ways to study the performance of an unsupervised model, so we take the opportunity to present additional results that provide more insight into the CRBM.

%%%
\begin{figure}[t!]
\includegraphics[width=6.5in]{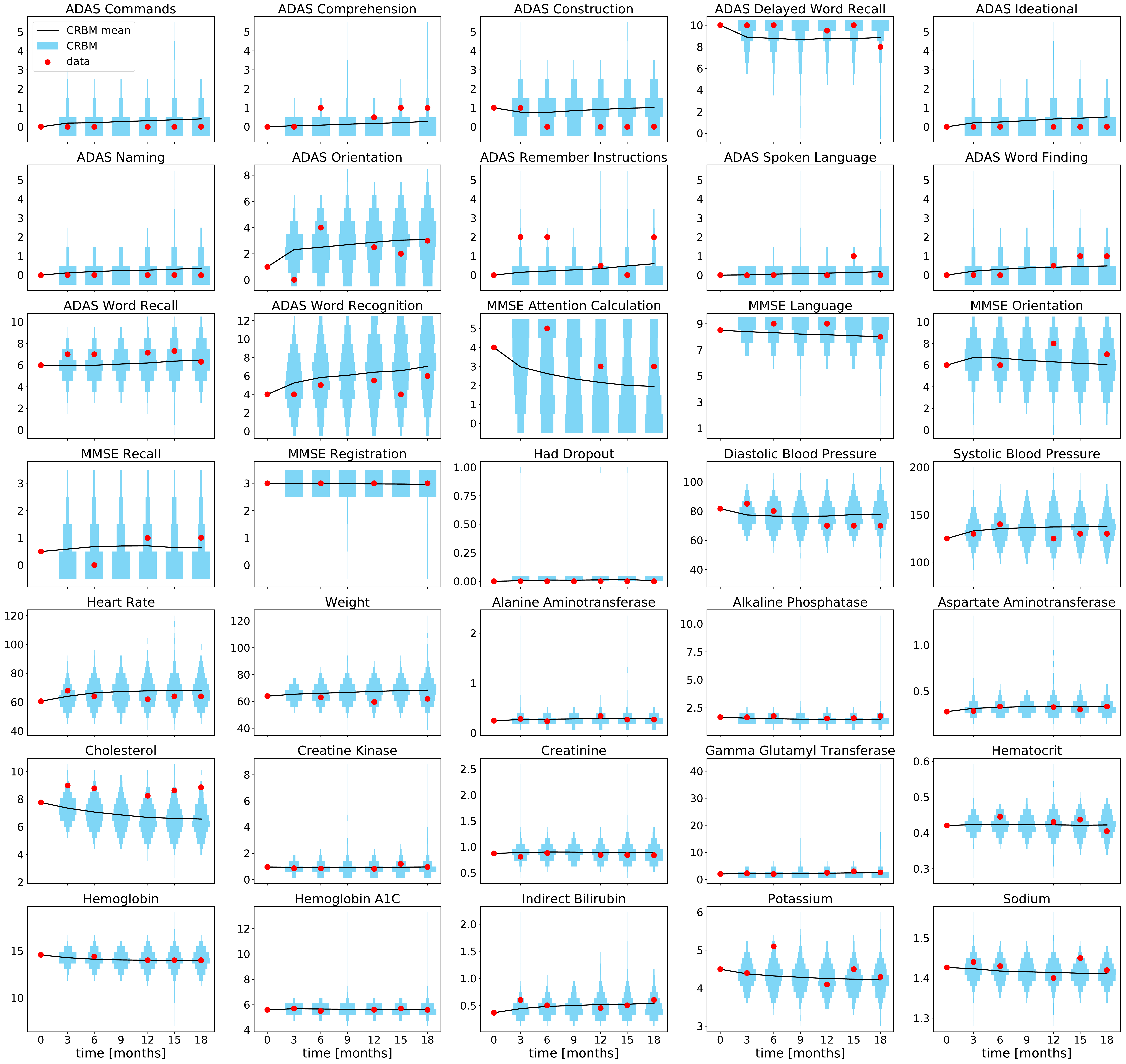} 
\caption{{\bf Stochastic simulations enable individual assessments of risk}.
Violin plots display the stochastic evolution of an MCI patient whose ADAS score change over 18 months suggests a conversion to AD. The width of the blue bars represents the probability computed using simulations from the CRBM, and the mean CRBM prediction is shown as the black line. The red dots show the actual observed values from the chosen patient. The CRBM was initialized with the observed values at baseline ($t=0$). Then, we repeatedly simulated 18-month trajectories and created histograms of each variable at every time point. The model predicts trends and imputes values when observations are missing for the patient (e.g., lab scores such as cholesterol and hematocrit). Units for these data are given in \Tab{non-cognitive-variables-specific}. 
\label{fig:risk}}
\end{figure}
%%%

%%%
\begin{figure}[p]
\includegraphics[width=4.5in]{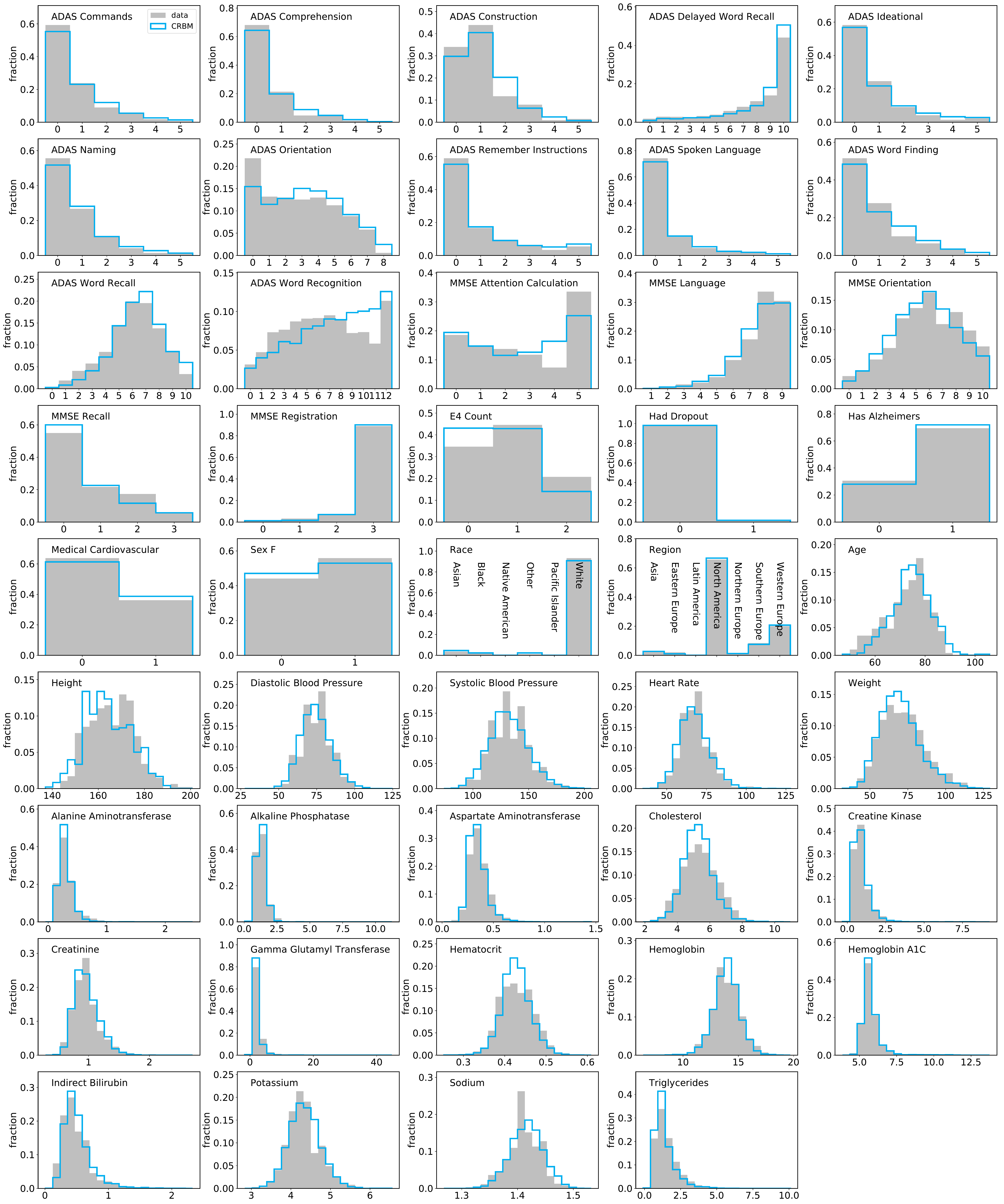} 
\caption{{\bf Marginal distributions in the generative mode for all variables.} The CRBM is used to generate 18-month patient trajectories, with the same number of virtual patients as the number of patients in the test group.  The marginal distributions for the patients and the CRBM are shown for all variables.  Note that some variables are constant over the virtual and real patients given the relatively small sample size (478 patients).
\label{fig:marginals_generative}}
\end{figure}
%%%

%%%
\begin{figure}[p]
\includegraphics[width=4.5in]{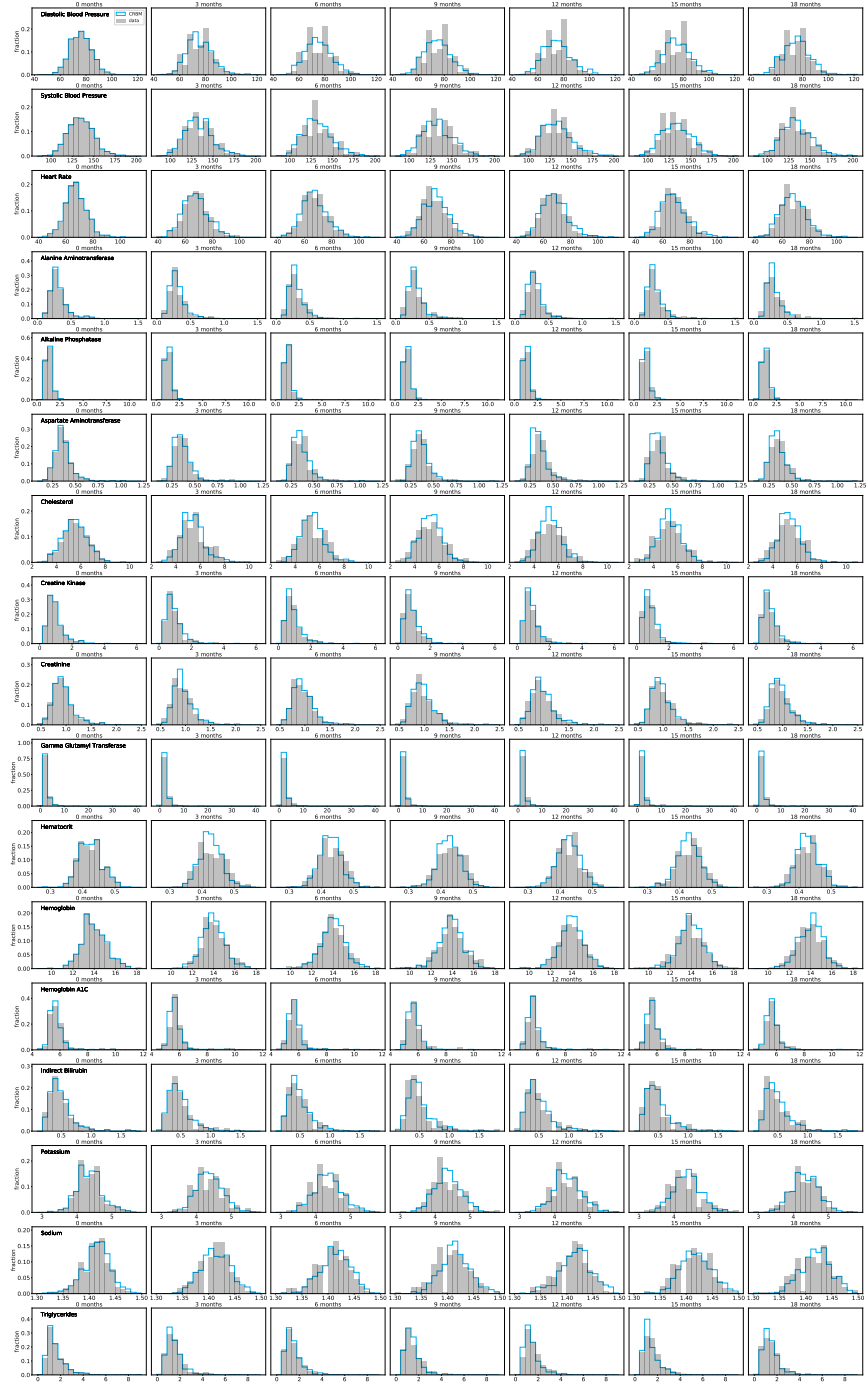} 
\caption{{\bf Marginal distributions conditioned on baseline for cognitive variables.} The CRBM is conditioned on the patient data at baseline, and one trajectory is simulated for each patient.  The marginal distributions for the patients and the CRBM are shown for temporal variables.  Because the CRBM is conditioned on the data at baseline, the 0-month distributions always match except when the data is missing values and the model performs imputation.
\label{fig:marginals_cognitive_conditioned}}
\end{figure}
%%%

%%%
\begin{figure}[p]
\includegraphics[width=4.5in]{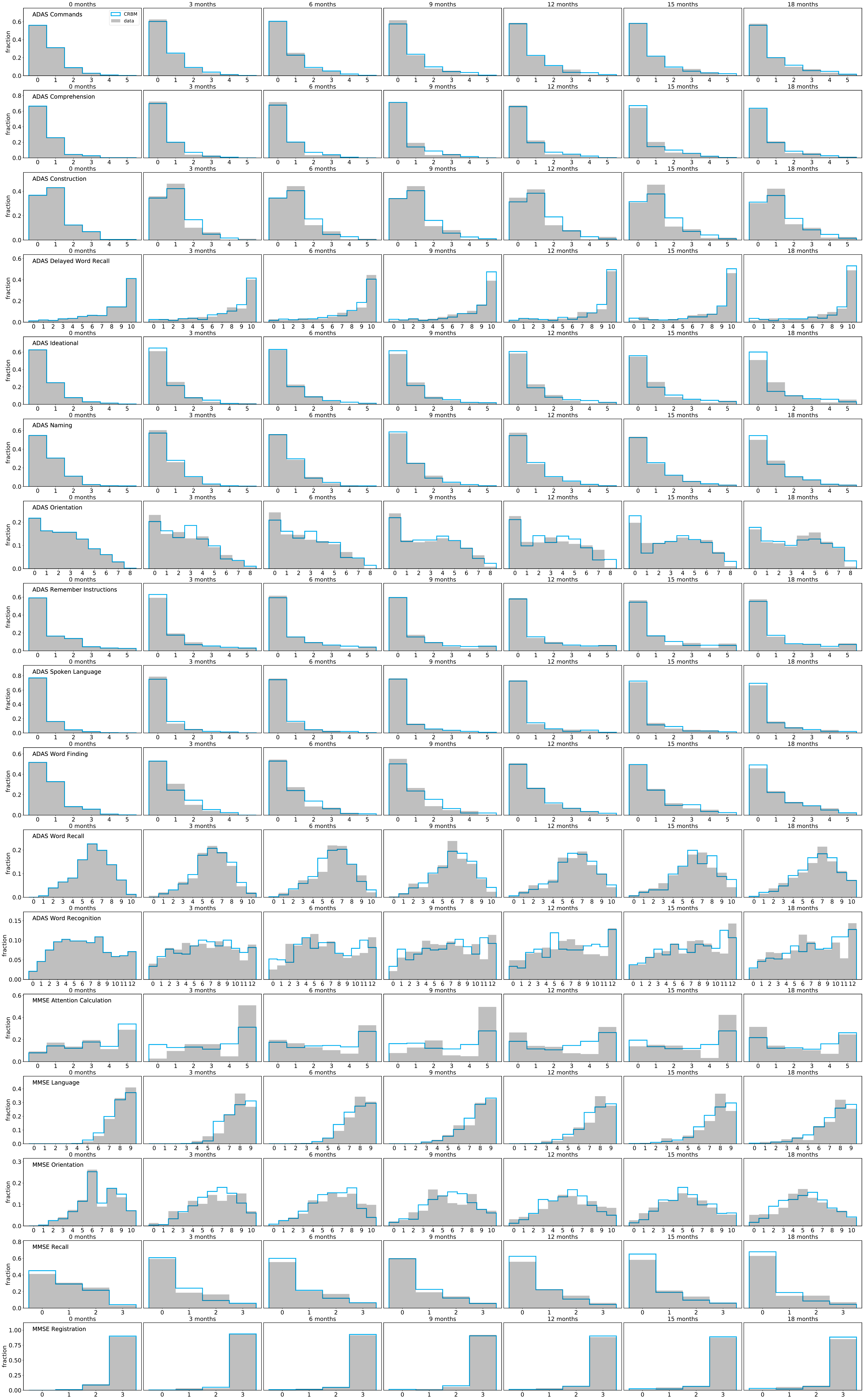} 
\caption{{\bf Marginal distributions conditioned on baseline for clinical variables.} The CRBM is conditioned on the patient data at baseline, and one trajectory is simulated for each patient.  The marginal distributions for the patients and the CRBM are shown for temporal variables.  Because the CRBM is conditioned on the data at baseline, the 0-month distributions always match except when the data is missing values and the model performs imputation.
\label{fig:marginals_clinical_conditioned}}
\end{figure}
%%%

%%%
\begin{figure}[t!]
\includegraphics[width=6.5in]{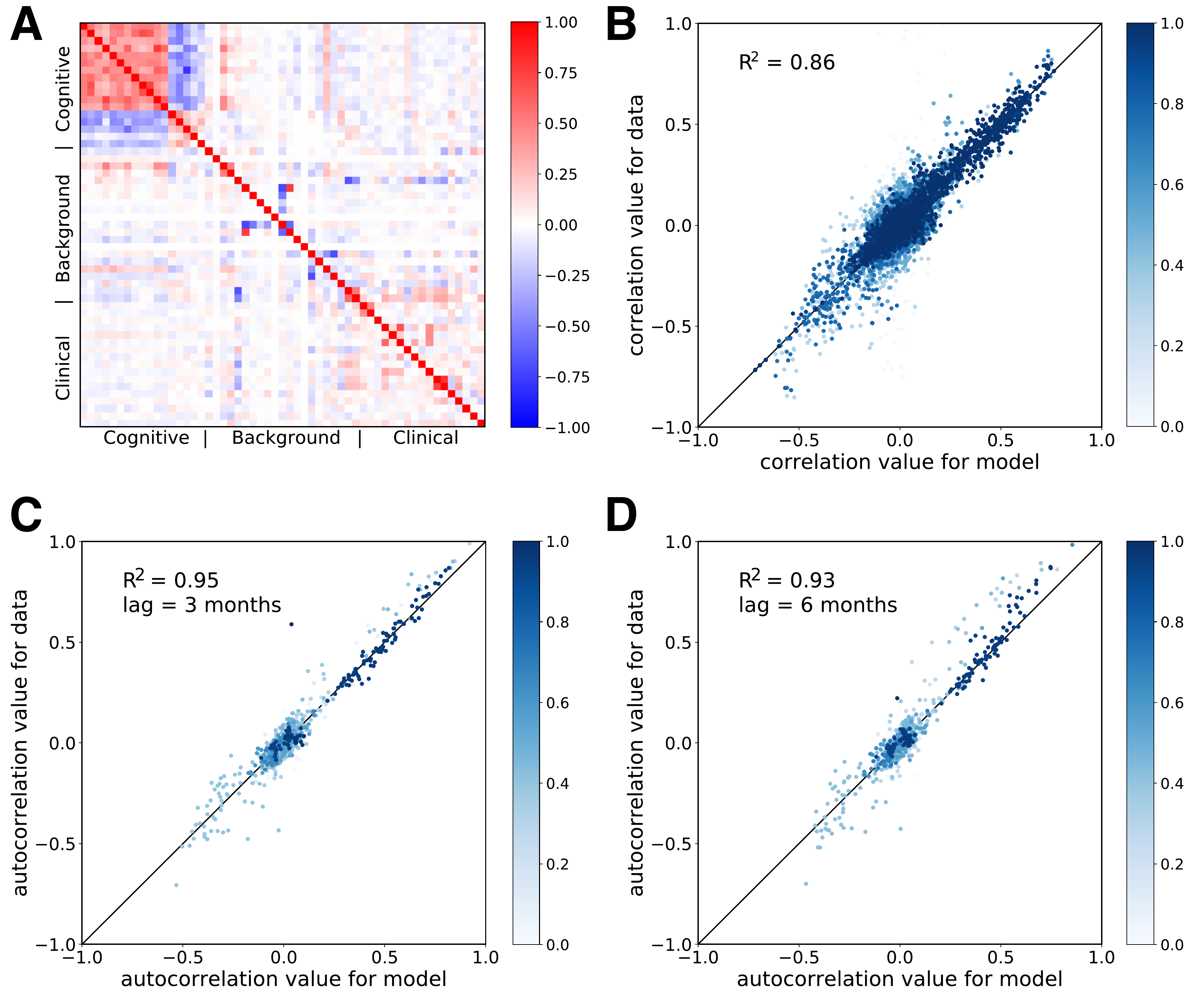} 
\caption{{\bf Goodness-of-fit}. A) Correlations between variables as predicted by the model (below the diagonal) conditioned on the data at baseline ($t=0$) and calculated from the data (above the diagonal). Components of the cognitive scores are strongly correlated with each other, but not with other clinical data. B) Scatterplot of observed vs predicted correlations for each time point, over all times. C) Scatterplot of observed vs predicted autocorrelations with time lag of 3 months. D) Scatterplot of observed vs predicted autocorrelations with time lag of 6 months. Color gradient in B-D represents the fraction of observations where the variables used to compute the correlation were present; lighter colors mean more of the data was missing.  This figure is a complement to Figure 2.
\label{fig:fit_conditioned}}
\end{figure}
%%%

%%%
\begin{figure}[t!]
\includegraphics[width=6.5in]{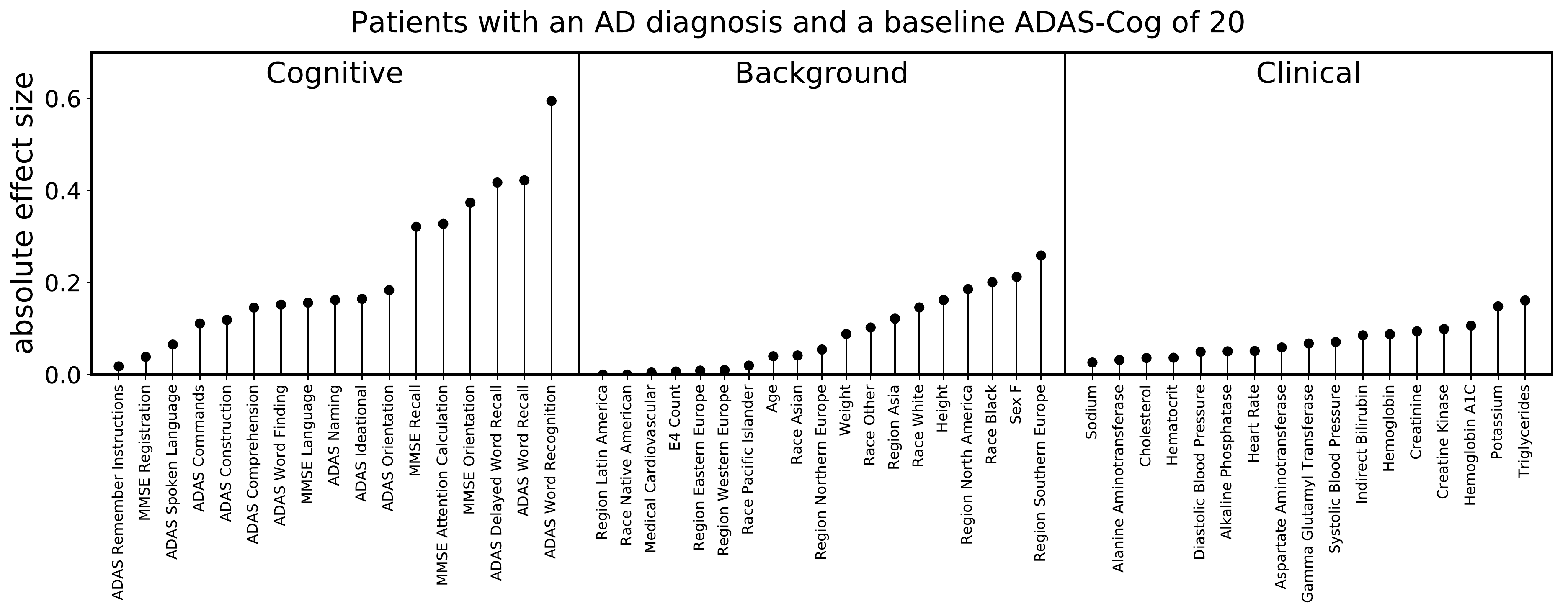} 
\caption{{\bf Using simulations to interpret prognostic signals for AD progression}. We created a simulated patient population with AD and an initial ADAS score of 20 (typical for AD), and simulated the evolution of each virtual patient for 18 months. The 5\% of virtual patients with the largest ADAS score increase were designated ``fast progressors'' and the bottom 5\% of patients with the smallest ADAS score increase were designated ``slow progressors''. Differences between the fast and slow progressors (the ``absolute effect size'') were quantified using the absolute value of Cohen's $d$-statistic.  This figure is a complement to Figure 4C.
\label{fig:interpretation_AD}}
\end{figure}
%%%

%%%
\begin{figure}[t!]
\includegraphics[width=6.5in]{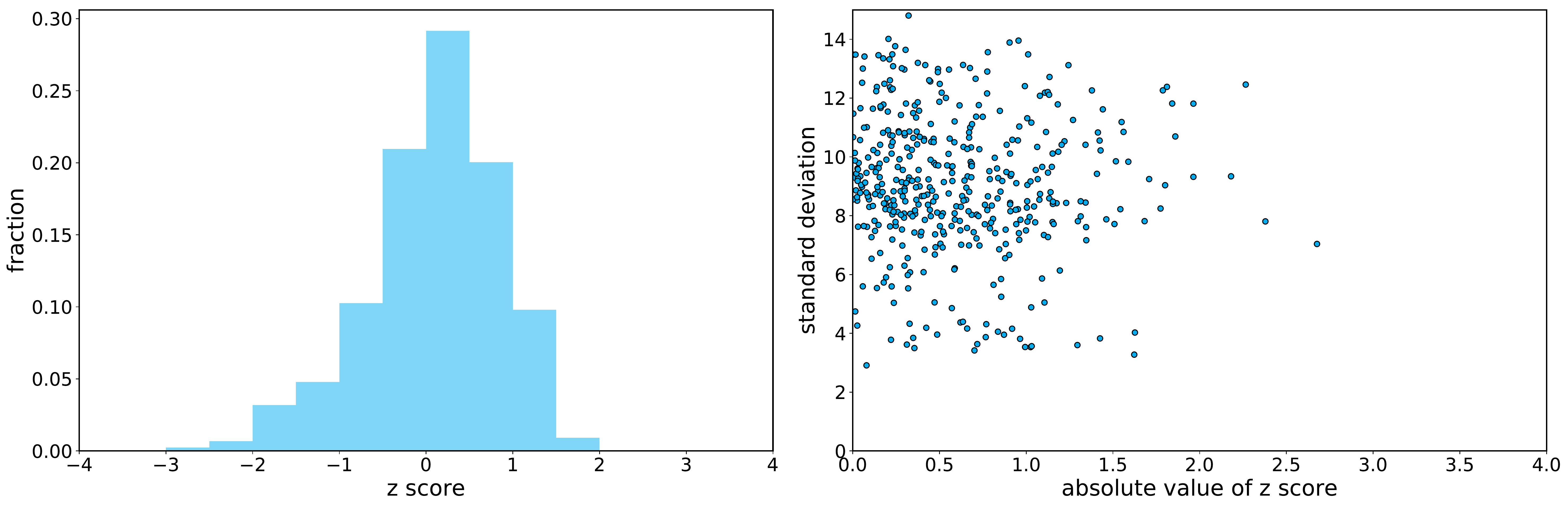} 
\caption{{\bf Confidence in ADAS score progression by the CRBM}.  For each patient with a valid ADAS score at baseline and 18 months, the CRBM is used to repeatedly simulate 18-month trajectories.  The mean of the ADAS score changes from these trajectories is the CRBM prediction, and the standard deviation is a measure of the CRBM confidence.  These values are used to compute a standard $z$ score for each patient by taking the difference between the CRBM prediction  and the true ADAS score change and dividing by the standard deviation of predictions.  These $z$ scores are 0-centered, tend to be fairly normally distributed (A), and do not correlate with the CRBM confidence (B).
\label{fig:score_change_z}}
\end{figure}
%%%

%%%
\begin{figure}[t!]
\includegraphics[width=6.5in]{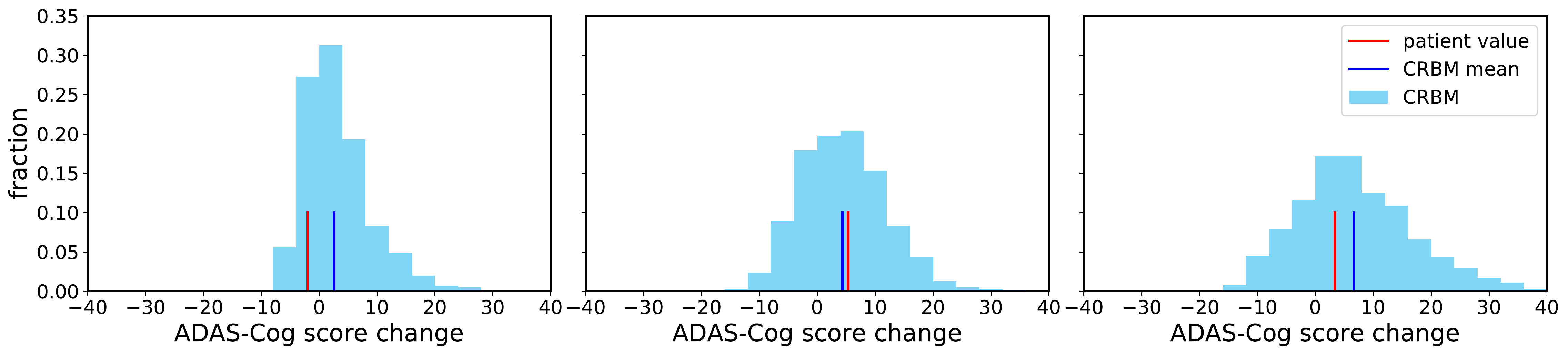} 
\caption{{\bf Predictions of ADAS score progression for example patients}.  Starting with baseline data for 3 example patients, the CRBM was used to predict the change in ADAS score over 18 months.  By repeatedly simulating trajectories for each patient, the CRBM provides a set of predictions per patient that forms a probability distribution.  The mean of this distribution is the CRBM prediction, which is compared with the true value of the ADAS score change for the patient.  The width of the distribution is a measure of the confidence of the CRBM prediction.
\label{fig:score_change_dists}}
\end{figure}
%%%

%%%
\begin{figure}[t!]
\includegraphics[width=6.5in]{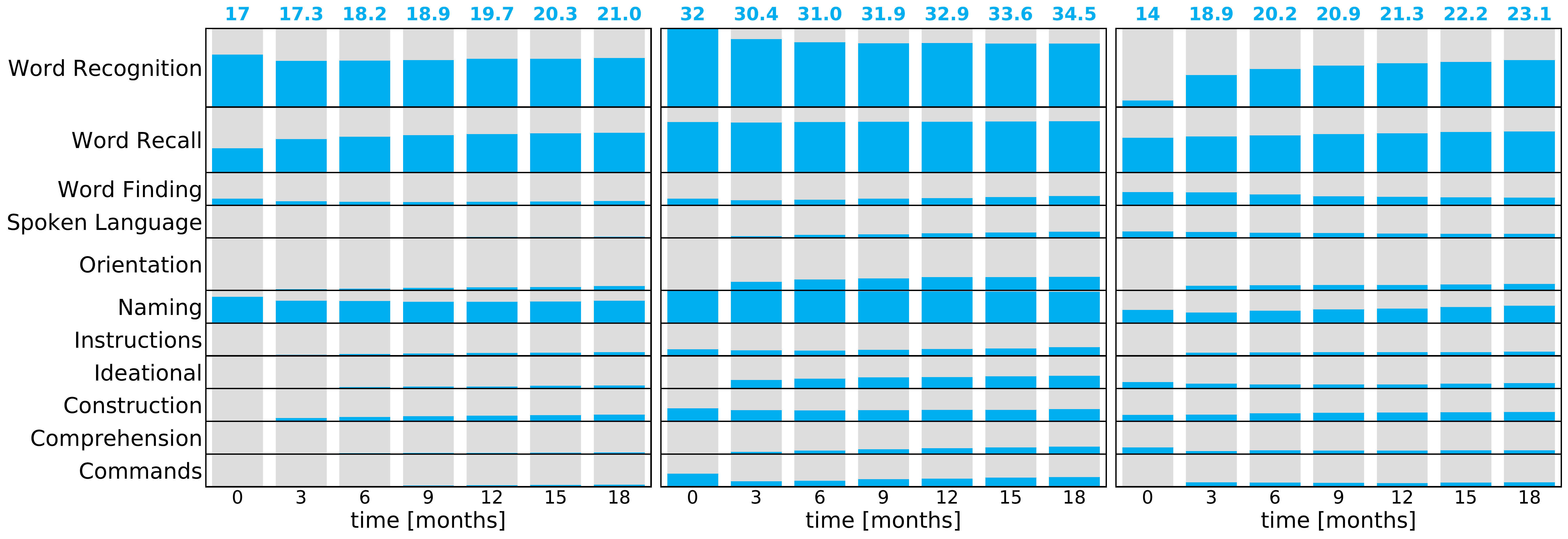} 
\caption{{\bf Expected evolution of ADAS score components for example patients}.  Starting with baseline data for 3 example patients, the CRBM was used to repeatedly simulate 18-month trajectories for each patient.  The mean value of each of the ADAS-Cog score components for each time point is shown as a blue bar, demonstrating the ability of the CRBM to simulate the granular ADAS score components.  The total mean ADAS score for each time point is shown at the top.
\label{fig:score_components}}
\end{figure}
%%%

%%%
\begin{figure}[thp!]
\includegraphics[width=3.2in]{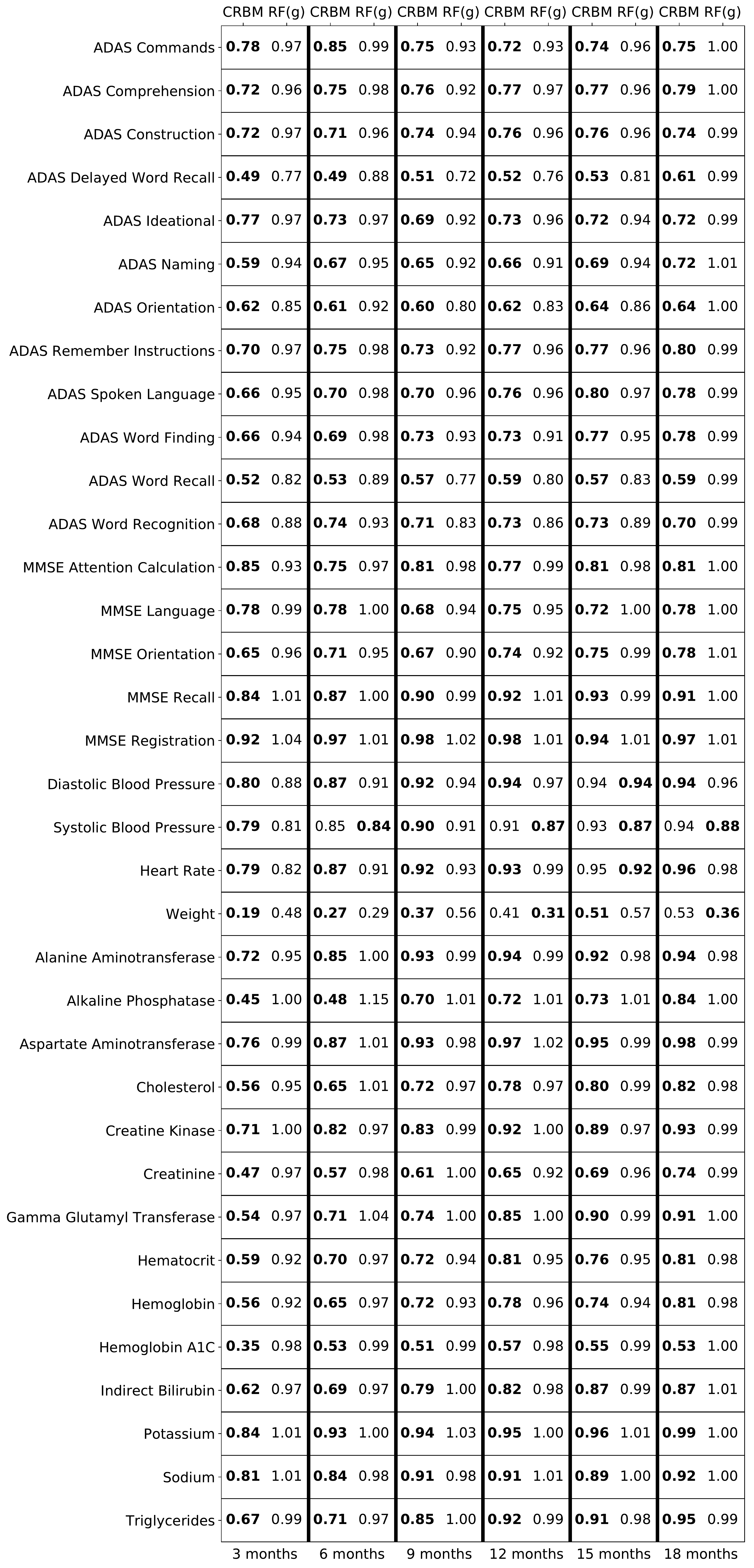} 
\caption{{\bf The model accurately forecasts across variables.} Relative errors of the model (CRBM) and a ``global" random forest (RF(g)) trained to predict the value of all variables at a single time point.  The root mean square (RMS) errors are scaled by the standard deviation of the data to be predicted.  Predictions are shown for every time-dependent variable except dropout.  At each time point and for each variable, the better of the random forest and CRBM predictions is shown in bold.  The CRBM strongly outperforms the global random forest.  This figure is a complement to Figure 3, where here the random forest is trained to predict all variables at a time point instead of having separate random forests for each variable.
\label{fig:trajectories_2}}
\end{figure}
%%%

\bibliography{camd}

\end{document}